\documentclass[10pt,twocolumn,letterpaper]{article}

\usepackage[pagenumbers]{cvpr} 
\usepackage{multirow}
\usepackage{comment}

\usepackage[dvipsnames]{xcolor}

\definecolor{cvprblue}{rgb}{0.21,0.49,0.74}
\usepackage[pagebackref,breaklinks,colorlinks,citecolor=cvprblue]{hyperref}

\title{ECRF: Entropy-Constrained Neural Radiance Fields Compression with Frequency Domain Optimization}

\author{
Soonbin Lee
\and
Fangwen Shu
\and
Yago Sánchez
\and
Thomas Schierl
\and
Cornelius Hellge
\and
Fraunhofer Heinrich-Hertz-Institute (HHI), Germany\\
{\tt\small \{first\_name\}.\{last\_name\}@hhi.fraunhofer.de}
}

\begin{document}
\maketitle
\begin{abstract}

Explicit feature-grid based NeRF models have shown promising results in terms of rendering quality and significant speed-up in training. However, these methods often require a significant amount of data to represent a single scene or object. In this work, we present a compression model that aims to minimize the entropy in the frequency domain in order to effectively reduce the data size. First, we propose using the discrete cosine transform (DCT) on the tensorial radiance fields to compress the feature-grid. This feature-grid is transformed into coefficients, which are then quantized and entropy encoded, following a similar approach to the traditional video coding pipeline. Furthermore, to achieve a higher level of sparsity, we propose using an entropy parameterization technique for the frequency domain, specifically for DCT coefficients of the feature-grid. Since the transformed coefficients are optimized during the training phase, the proposed model does not require any fine-tuning or additional information. Our model only requires a lightweight compression pipeline for encoding and decoding, making it easier to apply volumetric radiance field methods for real-world applications. Experimental results demonstrate that our proposed frequency domain entropy model can achieve superior compression performance across various datasets. The source code will be made publicly available.
\end{abstract}    
\section{Introduction}
\label{sec:intro}

Recent advancements in implicit neural representations (INRs)  have greatly influenced the research of 3D scene modeling and novel-view synthesis \cite{mildenhall2021nerf,sitzmann2020implicit,barron2021mip,tancik2020fourier}. This approach represents a complex volumetric scene as an implicit function that maps positional and directional information of sampled points to corresponding color and density values. This enables the rendering of photo-realistic novel views from desired viewpoints. Further developments have shown the capability to reconstruct various 3D scenes using images and corresponding camera poses, making radiance fields a promising method for representing the real 3D world \cite{zhang2020nerf++,barron2022mip,tewari2022advances}. However, these methods have a considerable computational overhead during training and inference, which is a significant bottleneck. This often results in several days of convergence time. 

\begin{figure}[t]
     \centering
     \begin{subfigure}[t]{\linewidth}
         \centering
         \includegraphics[width=\linewidth, height=2.85cm]{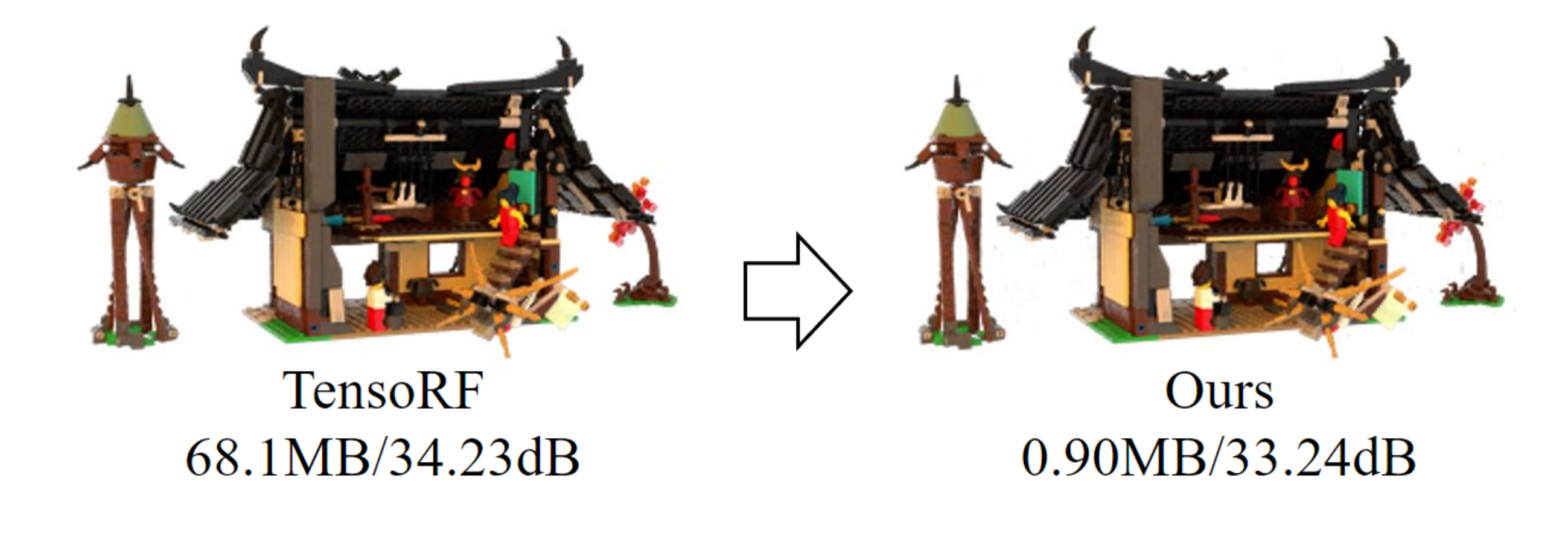}
         \caption{}
    \end{subfigure}
     \begin{subfigure}[t]{0.49\linewidth}
         \centering
         \includegraphics[width=\linewidth, height=3cm]{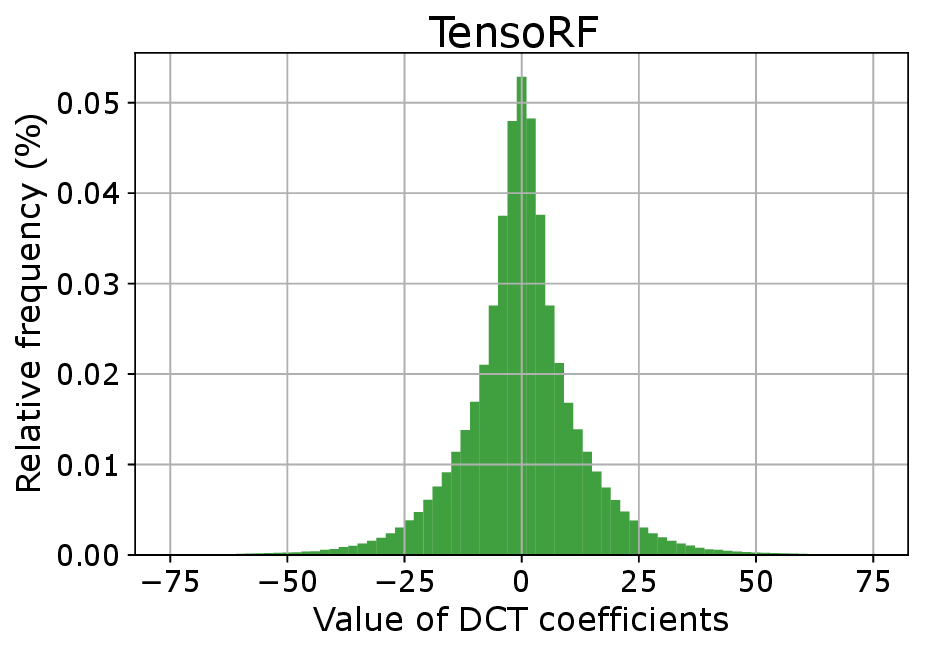}
         \caption{}
     \end{subfigure}
          \begin{subfigure}[t]{0.49\linewidth}
         \centering
         \includegraphics[width=\linewidth, height=3cm]{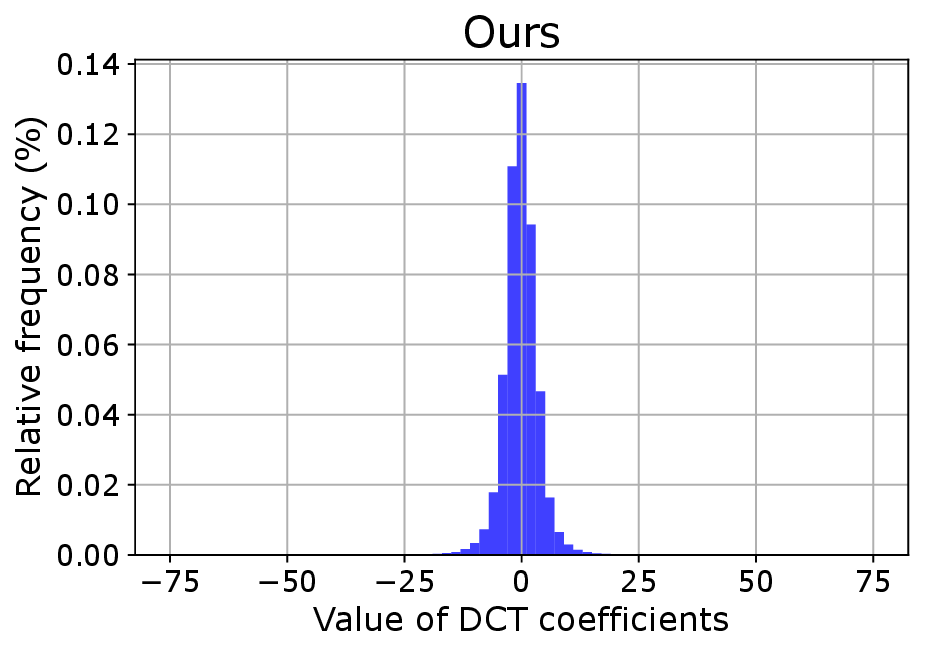}
         \caption{}
     \end{subfigure}

        \caption{\textbf{Reconstructed results} for the \textit{Blacksmith} scene from the Ray-Traced Multi-View (RTMV) dataset \cite{tremblay2022rtmv} are shown in (a). The frequency distribution of transformed coefficients is compared between the (b) TensoRF baseline and (c) the proposed model.}
        \label{fig:fig1}
\end{figure}

On the other hand, representing neural radiance fields using explicit grid-based structures has been shown to significantly improve training and inference efficiency \cite{fridovich2023k,sun2022direct, sun2022improved,chan2022efficient,muller2022instant}. These volumetric radiance field methods typically use grids to store learnable feature vectors and retrieve the color and density of a 3D point through trilinear interpolation. This can be achieved either without a neural network or with shallow MLPs, resulting in accelerated speed and high synthesis quality \cite{yu2021plenoctrees,yu2021plenoxels}. TensoRF is a parameter-efficient yet expressive method for decomposing dense 3D grids into a smaller set of parameters, such as matrices and vectors. This decomposition can significantly reduce the memory and computational requirements \cite{chen2022tensorf}. Similarly, by incorporating this feature-grid, the training and inference time of grid-based methods has been reduced from days to minutes \cite{sun2022improved,muller2022instant}. While these methods reduce the time complexity for training and inference in representing 3D scenes and objects, they have larger overall sizes compared to MLP-only methods. As a result, the use of grid-based volumetric representations inevitably leads to significant storage costs, which may restrict its practicality in real-world applications.

Therefore, in this work, we propose a new model called entropy-constrained radiance fields (ECRF) for compressing radiance fields. This optimization process leads to a compact feature representation in the frequency domain. As a result, the optimized radiance field can effectively reconstruct complex 3D scenes using a compact representation. Optimization leads to a sparse representation of transform coefficients, which allows the proposed model to outperform other compression models in terms of rate-distortion. As shown in Fig. \ref{fig:fig1}, the volumetric scene can be compressed by using compact transformed coefficients. The proposed model achieves a compression ratio of 75$\times$ by reducing the original size, while only resulting in a visual quality loss of 1dB for the given example. This paper presents experimental results to validate the effectiveness of the proposed model, which uses entropy-minimized frequency domain representations. We summarize our contributions as follows: 
\begin{itemize}
\item We propose a novel entropy-aware training scheme, which aims to compress data by minimizing log-likelihood with an entropy estimation network. During the training process, the optimization objective is to minimize the length of a bit sequence, which corresponds to the size of the entropy-coded bitstream.
\item We introduce the frequency-domain entropy parameterization. In order to leverage the sparsity of signals in the frequency domain, the original data is transformed into the frequency domain using block-wise discrete cosine transformation (DCT). To achieve high sparsity and effectively reduce the number of bits, we also apply regularization to the DCT coefficients during training. 
\item In the TensoRF-based compression models, we achieve state-of-the-art performance with a compression pipeline, which includes 8-bit quantization and entropy coding on the transformed coefficients.
\end{itemize}


\section{Related Work}
\label{sec:relatedwork}

\begin{figure*}[t]
 \centering 
 \includegraphics[width=\linewidth, height=5.5cm]{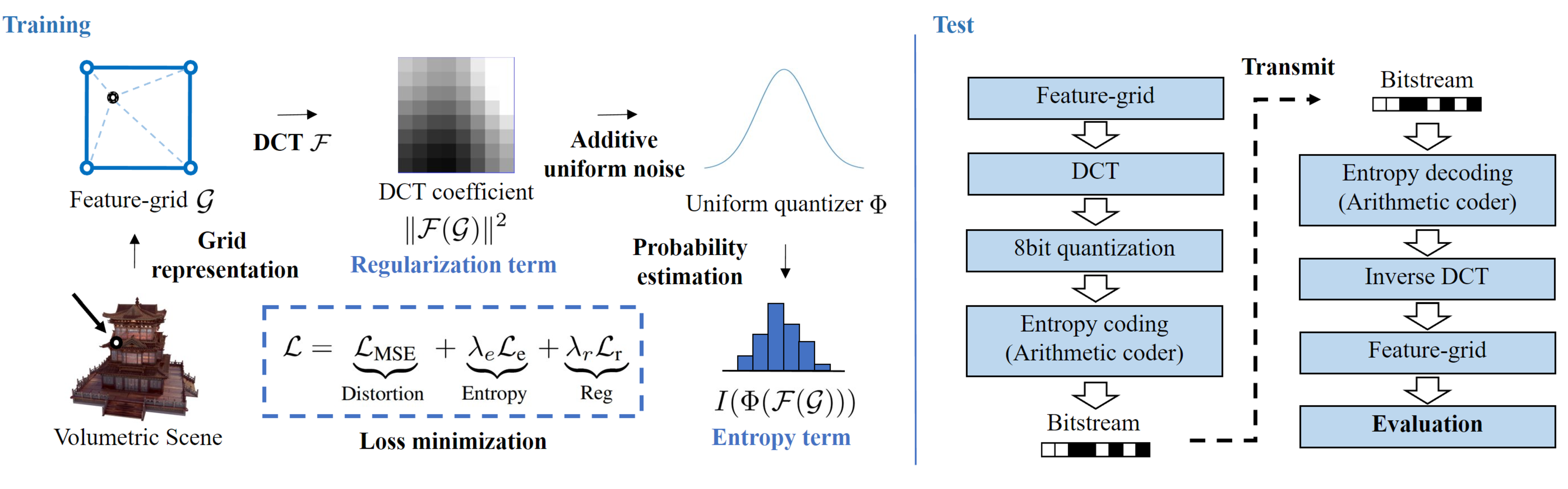}
 \caption{\textbf{Overall architecture of the proposed model.} After training with DCT representation as the optimization target, the feature-grid is entropy-coded in the frequency domain using the compression pipeline.}
 \label{fig:fig2}
\end{figure*}

\textbf{Radiance Field Compression.} To achieve real-time rendering with reasonable memory requirements, Instant-NGP proposed the use of multi-resolution voxel grids augmented with hash tables \cite{muller2022instant}. Another approach is to decompose 3D grids into lower-dimensional representations, such as vector-matrix and voxel grids \cite{chen2022tensorf,sun2022direct}. To reduce the size of these data structures, Variable Bitrate Neural Fields (VQAD) proposed using codebooks with vector quantization (VQ) to achieve variable bitrate and level of detail \cite{takikawa2022variable}. Similarly, VQRF proposes a compression framework pipeline that utilizes VQ with importance-based voxel pruning on the feature-grid \cite{li2023compressing}. VQRF compresion framework combined an 8-bit weight quantization with entropy coding to reduce the model size to 1MB, while still maintaining visual quality. CCNeRF suggests a low-rank approximation method for tensor decomposition \cite{tang2022compressible}, and Re:RF proposes simple pruning techniques to reduce the size of the feature-grid \cite{deng2023compressing}. 

The work most closely related to ours is the masked wavelet radiance field, which achieves competitive compression performance by using wavelet transformation and binarized masks for wavelet coefficients \cite{rho2023masked}. ReRF introduces DCT to compress the coefficients of the dynamic voxel grid using Huffman coding \cite{ReRF}. Similarly, TinyNeRF also incorporates DCT into the voxel grid and utilizes pruning and quantization-aware training to achieve high signal sparsity \cite{zhao2023tinynerf}. Another approach, Binary Radiance Fields (BiRF), utilizes binary encoding for parameter quantization, restricting values to either -1 or +1 to minimize feature vector storage requirements \cite{shin2023binary}. In this work, we demonstrate experimentally that using entropy parameterization can lead to more efficient representations, resulting in a smaller model size.



\textbf{Frequency Domain Transformation.} Several studies have examined the use of frequency-based parameterization to generate efficient neural field representations \cite{tancik2020fourier, wang2021neural, wu2023neural}. Fourier Plenoctree has demonstrated that applying the Fourier transform can enhance both the parameterization efficiency and training speed of the octree structure \cite{wang2022fourier}. However, these studies have focused on enhancing the expressiveness of 3D scenes through frequency representation, rather than fundamentally reducing model size. Meanwhile, conventional image and video codecs commonly utilize handcrafted techniques to reduce spatial redundancy. For example, the Joint Photographic Experts Group (JPEG) uses the wavelet transform to convert images from the pixel domain to the frequency domain \cite{marcellin2000overview,wallace1992jpeg}. In addition, High Efficiency Video Coding (HEVC) and Versatile Video Coding (VVC) employ DCT with varying block sizes \cite{sullivan2012overview,bross2021overview}. Inspired by this consideration, our work is to incorporate these video coding pipeline into the radiance field compression.

\textbf{Entropy Model for Learned Compression.} Several works for learned compression have been proposed to utilize the advanced capabilities of neural networks \cite{CGV-107,balle2020nonlinear,NEURIPS2020_8a50bae2}. The main idea behind neural compression is to use entropy models that are parameterized by neural networks \cite{balle2016end,balle2018variational,minnen2018joint}. This is achieved by establishing an entropy model that optimizes the size of its bitstream for latent features \cite{girish2023lilnetx,bird20213d,oktay2019scalable}. 


\section{Proposed Method}
\label{sec:proposed}


In this section, we introduce the entropy-constrained radiance field, a novel approach for efficient storage that employs entropy parameterization of transformed coefficients. Fig. \ref{fig:fig2} illustrates the overall scheme based on TensoRF \cite{chen2022tensorf} of our radiance field reconstruction. We will begin by explaining the parameterization of entropy using an entropy estimation network. Then, we will introduce a training approach that includes frequency domain parameterization and coefficient regularization. This approach aims to enhance data compression by leveraging the advantages of frequency domain representation. It is worth noting that our proposed model does not require any fine-tuning or additional storage cost (e.g., mask, index mapping). Once the feature-grid is trained with our loss function, the DCT is applied to transform the feature-grid into coefficients. These transformed coefficients are then quantized to 8-bit, following the same approach as other compression models \cite{li2023compressing, rho2023masked}. Finally, the quantized coefficients are encoded using an arithmetic coder to calculate the overall storage cost \cite{arithmetic}.

\subsection{Feature-Grid Neural Radiance Fields}
Neural Radiance Fields (NeRF) use an implicit function to represent scenes. This function maps spatial point $\mathbf{x}=(x, y, z)$ and view direction $\mathbf{d}=(\theta, \phi)$ to density $\sigma$ and color $\mathbf{c}$. By integrating the color $\mathbf{c}_i$ and density $\sigma_i$ of the spatial points $\mathbf{x}_i=\mathbf{o}+t_i \mathbf{d}$ sampled along a ray $\mathbf{r}$, we can estimate the RGB value $\hat{\boldsymbol{C}}(\boldsymbol{r})$ of the corresponding pixel:

\begin{equation}
\hat{\boldsymbol{C}}(\boldsymbol{r})=\sum_i^N T_i\left(1-\exp \left(-\sigma_i \delta_i\right)\right) \mathbf{c}_i
\end{equation}

In the above equation, $T_i=\exp \left(-\sum_{j=1}^{i-1} \sigma_i \delta_i\right)$, and $\delta_i$ represents the distance between adjacent samples. For efficient representation of 3D objects and scenes, recent studies have proposed the use of lower-dimensional grids, such as 2D planes or 1D lines \cite{chan2022efficient, chen2022tensorf}. TensoRF introduces a VM (vector-matrix) decomposition to reduce the memory usage caused by the large tensor $\mathcal{T} \in \mathbb{R}^{I \times J \times K}$ into low-rank matrices $\mathbf{M}$ and vectors $\mathbf{v}$ as follows:
\begin{equation}
\mathcal{T}=\sum_{r=1}^R \mathbf{v}_r^X \circ \mathbf{M}_r^{Y Z}+\mathbf{v}_r^Y \circ \mathbf{M}_r^{X Z}+\mathbf{v}_r^Z \circ \mathbf{M}_r^{X Y}
\end{equation}

Here, $\mathbf{v}_r^X \in \mathbb{R}^I$ denote the vector for $X$ axis, $\mathbf{M}_r^{Y Z} \in \mathbb{R}^{J \times K}$ is the matrix for $Y$ and $Z$ axes, $R$ is the rank, and the symbol $\circ$ is outer product. Hence, $\mathbf{v}_r^X \circ \mathbf{M}_r^{Y Z} \in \mathbb{R}^{I \times J \times K}$. TensoRF has shown its capability to decrease memory usage and training complexity by utilizing VM decomposition. However, this method results in total VM parameters that consume over 60MB of storage. In this work, we propose a compression model that reduces the size of the VM parameters by referring to them as a feature-grid. We will refer to the compression targets for all matrices $\mathbf{M}$ and vectors $\mathbf{v}$ as the feature-grid $\mathcal{G}$, and denote them as follows: 

\begin{equation}
\mathcal{G} = \left\{\mathbf{M}_\sigma, \mathbf{M}_c, \mathbf{v}_\sigma,\mathbf{v}_c\right\}
\end{equation}

A feature vector for a 3D point is calculated using trilinear interpolation from the feature-grid $\mathcal{G}$ and decoded by a small MLP to determine color and density values. 

\subsection{Entropy Parameterization of Feature-Grid}

We propose an entropy parameterization technique to the feature-grid $\mathcal{G}$ as an optimization target. By using this technique, the trained model minimizes its entropy $I$, resulting in a much shorter bit length:

\begin{equation}
I(\mathcal{G})=-\sum_{i=1}^N\log _2 P_{i}(\mathcal{G})
\end{equation}


To calculate the total entropy $I(\mathcal{G})$, we need to introduce a probability density function (PDF) model $P_{i}$ for $N$ samples $i \in\{1, \ldots, N\}$. However, estimating the PDF presents a significant challenge due to its non-differentiability. Because the derivatives of the naive rounding operation are almost zero everywhere, gradient descent becomes ineffective. To address this issue, the learned compression model proposed the use of additive uniform noise as an approximation for quantized functions \cite{balle2016end}. Similarly, we introduce a uniform quantization function $\Phi$ to obtain proxy gradients of the PDF:

\begin{equation}
\Phi(\theta)=\theta+u, u \sim \mathcal{U}\left(-\frac{1}{2}, \frac{1}{2}\right)
\end{equation}
where $\mathcal{U}\left(-\frac{1}{2}, \frac{1}{2}\right)$ is independent and identically distributed and uniformly distributed noise. This uniform quantization has been shown to be a good approximation for entropy when using the negative log-likelihood based on noisy discretization. Then, the discrete range is convolved with $P$ to obtain the discretized PDF:
\begin{equation}
\begin{aligned}
P(\Phi(\theta_{i}))) & =\int_{\Phi(\theta_{i})-\frac{1}{2}}^{\Phi(\theta_{i})+\frac{1}{2}} {P}(\theta_{i}) \mathrm{d} \theta_{i}
\\
& =P_{c}\left(\Phi(\theta_{i})+\frac{1}{2}\right)-P_{c}\left(\Phi(\theta_{i})-\frac{1}{2}\right)
\end{aligned}
\end{equation}

To simplify the calculation of PDF, \cite{balle2018variational} constructed a cumulative distribution function (CDF) $P_{c}$ with small MLPs. This CDF $P_{c}$ maps $\mathbb{R} \rightarrow[0,1]$ and it is a monotone increasing function that represents the CDF of the underlying PDF $P$. In this way, the calculation of $P_{c}$ simplifies the estimation of the PDF. Our proposed model only requires small MLPs to calculate $P_{c}$. After training is completed, these MLPs can be discarded. The detailed architecture of this entropy estimation network is described in the supplementary materials.

\subsection{Frequency Domain Parameterization}
The DCT is a commonly used method for Fourier-related transformations in image and video compression and it is known for its efficient energy compaction of spectral information. Hence, our proposed method focuses on signal compression by optimizing it in the frequency domain, rather than directly estimating entropy in the feature-grid. We estimate entropy in the frequency domain, which consists of transformed coefficients of the feature-grid. Specifically, the matrices $\mathbf{M}$ and vectors $\mathbf{v}$ are transformed into the frequency domain using the DCT.  
Given a feature-grid  $\mathcal{G} \in \mathbb{R}^{N \times C \times H \times W}$, where $N$ is the number of matrices and vectors, $C$ is the dimension of the feature channel, and $H \times W$ the shape of the feature-grid. The transformed coefficients $\mathcal{F(G)}$ after applying $3 \mathrm{D}$ DCT with $K_1$$\times$$K_2$$\times$$K_3$ block size can be defined as:
\begin{equation}
\begin{gathered}
\mathcal{F}(\mathcal{G})_{n, u, v, w}=\sum_{x=0}^{K_1-1} \sum_{y=0}^{K_2-1} \sum_{z=0}^{K_3-1} \mathcal{G}_{n, x, y, z} \\
\resizebox{0.48\textwidth}{!}{$
\cos \left[\frac{\pi}{K_1}\left(x+\frac{1}{2}\right) u\right] \cos \left[\frac{\pi}{K_2}\left(y+\frac{1}{2}\right) v\right] \cos \left[\frac{\pi}{K_3}\left(z+\frac{1}{2}\right) w\right]
$}
\end{gathered}
\end{equation}

\begin{figure}[t]
 \centering
      \begin{subfigure}[b]{0.48\linewidth}
         \centering
         \includegraphics[width=0.75\linewidth]{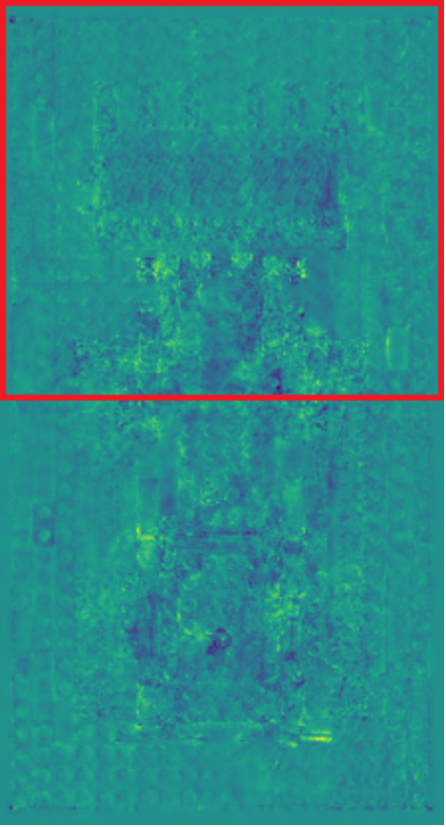}
         \caption{TensoRF}
     \end{subfigure}
     \vspace{0.01\linewidth}
          \begin{subfigure}[b]{0.48\linewidth}
         \centering
         \includegraphics[width=0.75\linewidth]{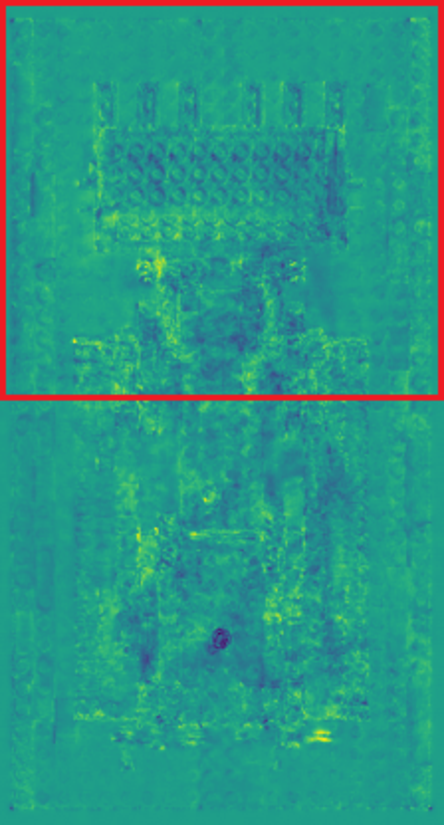}
         \caption{Ours}
     \end{subfigure}

      \begin{subfigure}[b]{0.48\linewidth}
         \centering
         \includegraphics[width=0.7\linewidth]{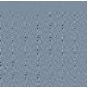}
         \caption{TensoRF}
     \end{subfigure}
     \vspace{0.01\linewidth}
          \begin{subfigure}[b]{0.48\linewidth}
         \centering
         \includegraphics[width=0.7\linewidth]{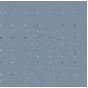}
         \caption{Ours}
     \end{subfigure}
 \caption{\textbf{Visualization of the first channel in the XY plane $\mathbf{M}_{c}^{XY}$} is shown for (a) TensoRF baseline and (b) our proposed model. The coefficients are transformed using a 2D DCT 32$\times$32 block $\mathcal{F}(\mathbf{M}_{c}^{XY})$ shown for (c) TensoRF baseline and (d) the proposed model. All results are based on 30k training iterations. 
 }
 \label{fig:fig3}
\end{figure}

\begin{table*}[t]
  \centering
  \resizebox{\textwidth}{!}{
\begin{tabular}{cccccccccc}
\hline
\multicolumn{1}{c|}{\multirow{5}{*}{Method}}                                          & \multicolumn{3}{c|}{Synthetic NeRF}                                                             & \multicolumn{3}{c|}{Synthetic NSVF}                                                             & \multicolumn{3}{c}{Tanks and Temples}                                      \\
\multicolumn{1}{c|}{}                                                                 & \multirow{4}{*}{Size (MB)↓} & \multirow{4}{*}{PSNR↑} & \multicolumn{1}{c|}{\multirow{4}{*}{SSIM↑}} & \multirow{4}{*}{Size (MB)↓} & \multirow{4}{*}{PSNR↑} & \multicolumn{1}{c|}{\multirow{4}{*}{SSIM↑}} & \multirow{4}{*}{Size (MB)↓} & \multirow{4}{*}{PSNR↑} & \multirow{4}{*}{SSIM↑} \\
\multicolumn{1}{c|}{}                                                                 &                            &                       & \multicolumn{1}{c|}{}                      &                            &                       & \multicolumn{1}{c|}{}                      &                            &                       &                       \\
\multicolumn{1}{c|}{}                                                                 &                            &                       & \multicolumn{1}{c|}{}                      &                            &                       & \multicolumn{1}{c|}{}                      &                            &                       &                       \\
\multicolumn{1}{c|}{}                                                                 &                            &                       & \multicolumn{1}{c|}{}                      &                            &                       & \multicolumn{1}{c|}{}                      &                            &                       &                       \\ \hline
TensoRF-VM \cite{chen2022tensorf}                                                                       & 71.3                       & 33.15                 & 0.962                                      & 71.8                       & 36.70                 & 0.984                                      & 72.0                       & 28.58                 & 0.922                 \\
Instant-NGP \cite{muller2022instant}                                                                      & 39.5                       & 33.08                 & 0.960                                      & 39.5                       & 36.04                 & 0.981                                      & 39.5                       & 28.85                & 0.924                 \\
TensoRF-CP \cite{chen2022tensorf}                                                                       & 3.9                        & 31.66                 & 0.950                                      & 3.9                        & 34.48                 & 0.971                                      & 4.0                        & 27.59                 & 0.897                 \\ \hline
Re:TensoRF High \cite{deng2023compressing}
& 7.9                          & 32.81                 & 0.956                                      & 8.5                          & 36.14                & 0.978                                      & 6.7                          & 28.24                 & 0.907                \\
TinyNeRF (4MB) \cite{zhao2023tinynerf}
& 4                          & 31.90                 & 0.956                                      & 4                          & 34.88                 & 0.975                                      & 4                          & 28.32                 & 0.910                 \\ TinyNeRF (2MB) \cite{zhao2023tinynerf}
& 2                          & 31.72                 & 0.954                                      & 2                          & 34.62                 & 0.968                                      & 2                          & 28.19                 & 0.908                 \\
VQ-TensoRF \cite{li2023compressing}                                                                       & 3.5                        & 32.88                 & 0.960                                      & 4.2                        & 36.04                 & 0.979                                      & 3.4                        & 28.25                 & 0.913                 \\
VQ-DVGO \cite{li2023compressing}                                                                          & 1.4                        & 31.77                 & 0.954                                      & 1.3                        & 34.72                 & 0.974                                      & 1.4                        & 28.26                 & 0.909                 \\ \hline
BiRF (Rate-3) \cite{shin2023binary}
& 2.8                        & \textbf{33.26}                 & 0.961                                      & 2.9                        & 36.17                 & 0.983                                      & 2.9                        & \textbf{28.62}                 & 0.924                 \\
\begin{tabular}[c]{@{}c@{}}Masked (Rate-3) \cite{rho2023masked}\end{tabular} & 2.4                        & 32.38                 &   0.958                                         & 2.7                        & 35.57                 &  0.976                                          & 2.4                        & 28.27                 &     0.915                  \\
\textbf{Ours (Rate-3)}                                                                               & 2.5                        & 33.05                 &    0.961                                        &  2.6                          & \textbf{36.22}                      &    0.983                                        &  2.5                          &  28.41                     &  0.921                     \\ \hline

BiRF (Rate-2) \cite{shin2023binary}
& 1.4                        & 32.64                 & 0.959                                      & 1.5                        & 35.40                 & 0.976                                      & 1.5                        & \textbf{28.44}                 & 0.916                 \\
\begin{tabular}[c]{@{}c@{}}Masked (Rate-2) \cite{rho2023masked}\end{tabular}   & 1.5                        & 32.23                 &   0.958                                         & 1.6                        & 35.33                 &   0.976                                         & 1.7                        & 28.01                 &     0.904                  \\
\textbf{Ours (Rate-2)}                                                                               & 1.3                        & \textbf{32.72}                 &    0.960                                        &    1.4                        &  \textbf{35.73}                     & 0.978                                           &   1.4                         &    28.31                  &  0.915                     \\ \hline

BiRF (Rate-1) \cite{shin2023binary}
& 0.7                        & 31.53                 & 0.949                                      & 0.8                        & 34.26                 & 0.971                                      & 0.8                        & 28.02                 & 0.906                 \\
\begin{tabular}[c]{@{}c@{}}Masked (Rate-1) \cite{rho2023masked}\end{tabular}   & 0.8                        & 31.95                 &      0.956                                      & 0.9                        & 34.67                 &  0.974                                          & 0.9                        & 27.77                 &     0.901                  \\
\textbf{Ours (Rate-1)}                                                                               & 0.8                        & \textbf{32.20}                 &     0.958                                       & 0.8                        & \textbf{35.00}                 &  0.975                                          &    0.8                        &   \textbf{28.14}                     &  0.908                     \\ \hline
\end{tabular}
}
\caption{\textbf{Quantitative results} of model size and reconstruction quality are presented for the Synthetic NeRF, NSVF dataset, and Tanks and Temples dataset. To provide a more detailed comparison, we report the performance at three rate levels: \{0.8, 1.4, 2.5\}MB. Note that these models use different experimental configurations for each rate level, as described in Sec \ref{sec:sec4.2}. The results of BiRF, TinyNeRF and
Re:TensoRF are directly adopted from the original paper.}
\label{tab:tab1}
\end{table*}

\noindent where ${\mathcal{F(G)}} \in \mathbb{R}^{N \times C \times H \times W}$ is the transformed coefficients, $[x, y, z]$ and $[u, v, w]$ are block indices for $\{0,1,2,\ldots,K_i-$1$\}$. Fig. \ref{fig:fig3} provides a visualization of the XY plane tensor $\mathbf{M}_{c}^{XY}$, using a 2D DCT with block size 32$\times$32 for \textit{Lego} scene. The signal on the tensor plane projected onto the XY-axis can be seen in the figure. During training, our proposed method can use frequency domain transformation to generate optimized representations for all transformed coefficients. After training with the entropy minimization, it is shown that the transformed coefficients $\mathcal{F}(\mathbf{M}_{c}^{XY})$ can restore the original tensor signal with a more sparse representation.  For each volumetric scene, our proposed model jointly optimizes the entropy of the transformed coefficients, resulting in a sparse representation. The DCT operation is differentiable, allowing for backpropagation with entropy parameterization. 

\pagebreak
Finally, our proposed model uses a differentiable entropy of transformed coefficients $\mathcal{F(G)}$ as a loss function $\mathcal{L}_{\text {e}}$:

\begin{equation}
\mathcal{L}_{\text {e}}= I(\Phi(\mathcal{F({\mathcal{G}})))}
\end{equation}
Additionally, we apply a coefficient regularization term. This regularization term is expected to have an impact on the quantization and rounding of the DCT coefficients, similar to the compression pipeline used in traditional video codecs. Experimental results have shown that this term effectively improves sparsity, leading to a smaller bitstream size. Specifically, we apply $\mathcal{L}_2$ norm regularization by squaring all values of the transformed coefficients for the following optimization objective:

\begin{equation}
\mathcal{L}_{\text {r}}=\|\mathcal{F(\mathcal{G})}\|^2
\end{equation}

To establish a rate-distortion trade-off, we consider the mean squared error (MSE) of the RGB value as the measure of distortion:

\begin{equation}
{\mathcal{L}_{\text {MSE}}}=\frac{1}{|\mathcal{R}|} \sum_{\boldsymbol{r} \in \mathcal{R}}\|\hat{\boldsymbol{C}}(\boldsymbol{r})-\boldsymbol{C}(\boldsymbol{r})\|^2
\end{equation}
where $\mathcal{R}$ is the set of camera rays sampled in a mini-batch, and $\boldsymbol{C}(\boldsymbol{r})$ is the ground-truth pixel value corresponding to the camera ray $r$. 
Thus, the overall loss function is a combination of the distortion loss and entropy parameterizations of the transformed coefficients with the regularization term:

\begin{equation}
\mathcal{L}=\underbrace{\mathcal{L}_{\text {MSE}}}_{\text {Distortion }}+\underbrace{\lambda_{e} \mathcal{L}_{\text {e}}}_{\text {Entropy }} + \underbrace{\lambda_{r}\mathcal{L}_{\text {r }}}_{\text {Reg}}
\end{equation}

Our proposed model incorporates this loss function that minimizes entropy in the transformed representation. It results in a relatively sparse discrete representation of the feature-grid, significantly reducing the size of the compressed model when entropy coding is performed. By adjusting the parameter $\lambda_{e}$ in different experiments, this model can explore the trade-off between compressed size and reconstructed visual quality. 

\begin{figure*}[t]
 \centering 
 \includegraphics[width=0.9\linewidth, height=10cm]{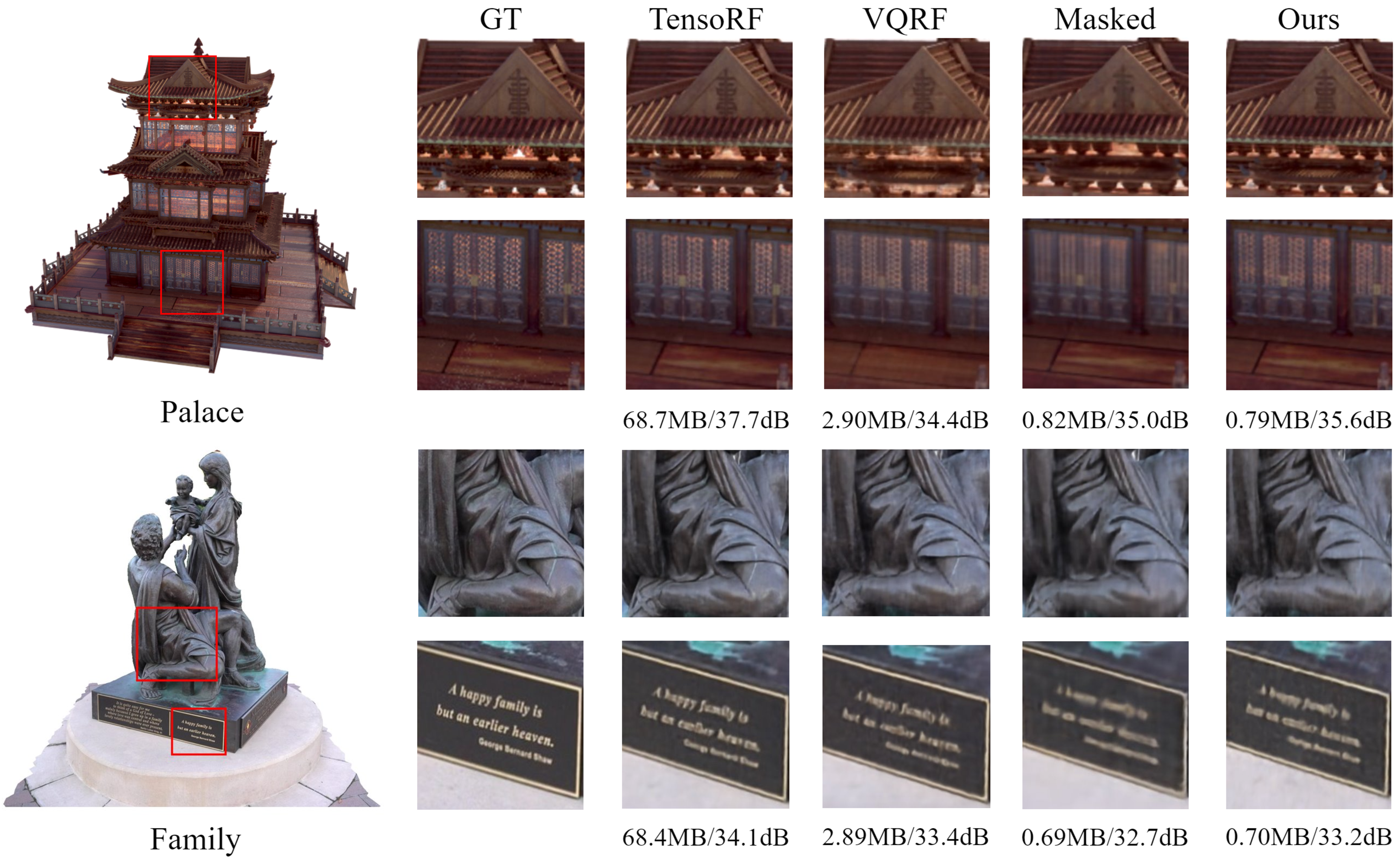}
 \caption{\textbf{Qualitative results} for visual comparison are provided for each scene of the Synthetic NSVF and Tanks and Temple datasets using TensoRF-based compression models. Each subfigure displays the storage size and PSNR.}
 \label{fig:fig4}
\end{figure*}


To render the original scene, the compression pipeline described above only needs one inverse DCT and entropy decoding for the transformed coefficients. Therefore, the additional complexity costs to the original feature-grid model are negligible. For TensoRF, a shallow decoder MLP is used to predict color and density from the trilinear interpolated feature-grid. The total storage required for a decoder MLP is approximately 0.14 MB, according to the default experimental configurations of TensoRF. In our quantitative results presented in Tab. \ref{tab:tab1}, we report the model size including the decoder MLP, and do not apply any compression techniques to it.

\section{Experiments and Results}
\label{sec:results}

We present a comprehensive evaluation of the proposed compression model for tensorial radiance fields. First, we describe our configurations and the datasets used. Next, we compare our approach with previous and concurrent works. Finally, we present ablation studies for each component and parameter.

\subsection{Experimental Configurations}
We follow the experimental configurations of TensoRF, which includes 30k iterations of training and a minibatch size of 4096 rays, along with other configurations \cite{chen2022tensorf}. TensoRF uses a coarse-to-fine training strategy that dynamically adjusts the tensor size and alpha mask (occupancy) to achieve accurate rendering. During training, the process of tensor upsampling and alpha mask updating has the potential to disrupt the accurate estimation of entropy. To update the alpha mask, we use a slightly different version from the masked wavelet model \cite{rho2023masked}, which is at the $\{2000, 4000, 6000, 11000, 16000\}$ iterations. The masked wavelet model utilizes a wider range of intervals, but this configuration does not have a significant impact on our model. Therefore, the loss function of our proposed model is applied starting from the 16k iteration, when the size of the feature-grid is completely fixed. It has been observed that the proposed model converges sufficiently under this condition.
\begin{table}[t]
  \centering
  \resizebox{0.5\textwidth}{!}{
\begin{tabular}{c|ccc}
\hline

Method          & Size (MB)↓               & PSNR (dB)↑           & SSIM↑                  \\ \hline
TensoRF-VM \cite{chen2022tensorf}      & $75.3 \mathrm{MB}$                       & 33.67                &  0.971       \\
VQRF \cite{li2023compressing}           &      $2.18 \mathrm{MB}$                    &      30.60                &  0.949                       \\
NGLOD-NeRF \cite{takikawa2021nglod}
& $20 \mathrm{MB}$ & 32.72 & 0.970 \\ \hline
VQAD (6bw) \cite{takikawa2022variable}
& $0.49 \mathrm{MB}$       & 30.76                & 0.956                 \\
\textbf{Ours ($\lambda_{e}=2e^{-10}$)}            & $0.41 \mathrm{MB}$       & \textbf{31.40}                &   \textbf{0.961}                     \\
VQAD (4bw) \cite{takikawa2022variable}
& $0.33 \mathrm{MB}$       & 30.09                & 0.948                 \\
\textbf{Ours ($\lambda_{e}=5e^{-10}$)}            & $0.29 \mathrm{MB}$       & \textbf{30.33}                &   \textbf{0.949}                     \\
 \hline
\end{tabular}
}
\caption{\textbf{Quantitative results} for 10 brick scenes from RTMV dataset at a 400$\times$400 resolution. The results of VQAD and NGLOD-NeRF are directly adopted from the original paper.}
\label{tab:tab2}
\end{table}

We then compare our model with other compression models using various datasets, including Synthetic NeRF and Neural Sparse Voxel Fields (NSVF) for 800$\times$800 resolution and Tanks And Temples (T\&T) for 1920$\times$1080 resolution \cite{mildenhall2021nerf,liu2020neural,tanks}. In addition, 10 brick scenes from the Ray-Traced Multi-View (RTMV) dataset are used for comparison with VQAD at a 400$\times$400, which matches the resolution used in the VQAD \cite{tremblay2022rtmv, takikawa2022variable}. The proposed model is based on TensoRF VM-192 and adjusts the parameter $\lambda_{e}$ to control the bitrate. Regarding the DCT block size, this paper chooses a 3D block size of 16$\times$16$\times$16 for matrices $\mathbf{M}$, and a 2D block size of 8$\times$8 for vectors $\mathbf{v}$. More details are presented in section \ref{sec:sec4.3}.

\begin{figure}[t]
 \centering
         \centering
         \includegraphics[width=\linewidth]{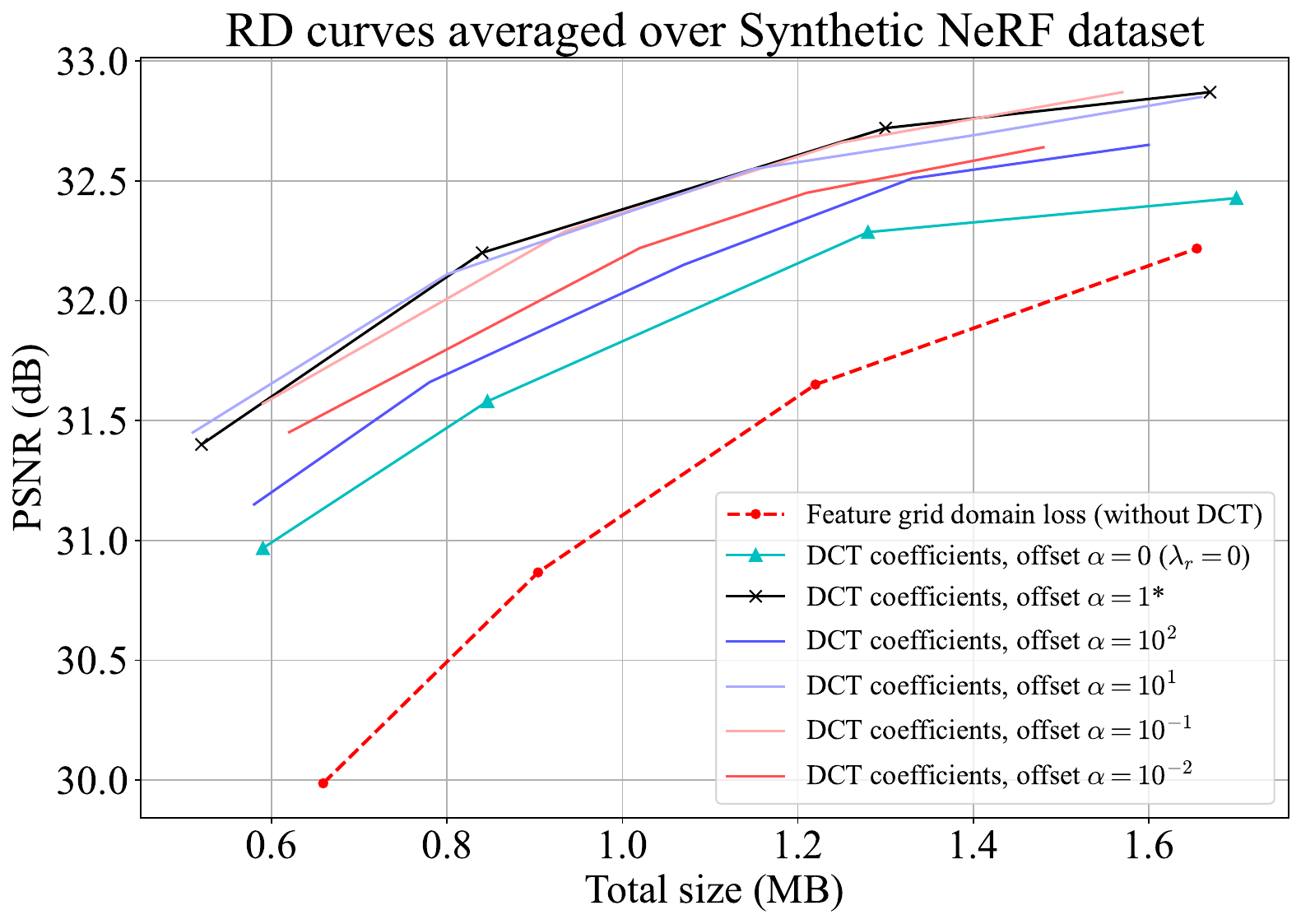}
         \caption{\textbf{Ablation study: the impact of the regularization term.} The rate-distortion (RD) curves demonstrate the effect of varying the regularization factor $\lambda_{r}$ with offset $\alpha$, where * indicates the default configuration. 
         }
     \label{fig:fig5}
\end{figure}

\subsection{Quantitative Result}
\label{sec:sec4.2}
For a fair comparison, different models are categorized into grid-based models and compression models in Tab. \ref{tab:tab1}. In this table, 'BiRF' indicates binary radiance fields that is based on Instant-NGP with binary encoding \cite{shin2023binary}, and 'Masked' represents the masked wavelet model that is based on the TensoRF with masking for wavelet coefficients \cite{rho2023masked}. To compare the performance of compression models, we assess the visual quality at three different rate levels: about 0.8MB (Rate-1), 1.4MB (Rate-2), and 2.5MB (Rate-3). In the low bitrate range (Rate-1), our proposed model achieved the highest performance across all datasets compared to the other compression models. We also present the performance of the masked wavelet model based on the configuration of VM-192 for Rate-1. It is important to note that the proposed model is based on the VM-192 configuration for all bitrate levels. This ensures a fair comparison with other TensoRF-based compression models.

In the middle bitrate range (Rate-2), BiRF achieves 0.13dB better visual quality than ours on the T\&T dataset, with a slightly larger size. It is worth noting that BiRF is based on Instant-NGP, which demonstrates superior quality on the T\&T dataset. Additionally, we report the performance of the masked wavelet model on VM-384 to achieve the better results at Rate-2 and -3. In VM-384, the tensor channel size is doubled from the default configuration to achieve the better results. Increasing the number of channels can increase the model capacity, but this configuration also leads to higher computational costs for both training and testing.

In the high bitrate range (Rate-3), our proposed model demonstrates better performance on the NSVF dataset, while BiRF shows better quality in the other two datasets. The base model clearly affects the compression performance results. Moreover, BiRF increases the number of feature dimensions $F$ from 2 to 4 at Rate-3, which is twice the default configuration. As a result, BiRF is reported to have high performance, even surpassing the TensoRF and Instant-NGP baseline in terms of visual quality.

The rendering results for visual comparison of each compression model are shown in Fig. \ref{fig:fig4}, where we illustrate that our proposed model can maintain better visual quality even in smaller sizes. Tab. \ref{tab:tab2} presents the comparison results between our proposed model and VQAD. VQAD employs vector quantization of the feature-grid by learning an integer codebook. Ours demonstrates better compression performance on the RTMV dataset.

\subsection{Ablation Study}
\label{sec:sec4.3}
\begin{table}
  \centering
  \resizebox{0.5\textwidth}{!}{
\begin{tabular}{c|c|ccc}
\hline
DCT block         & Total block size & Size (MB)↓ & PSNR (dB)↑ \\ \hline
2D DCT (1$\times$8$\times$8)     & $T=64$            &   0.60/1.05        &   29.86/32.22          \\
2D DCT (1$\times$16$\times$16)   & $T=256$             &  0.63/1.08         &  31.81/33.04           \\
2D DCT (1$\times$32$\times$32)  & $T=1024$            &   0.69/1.08        &   33.21/33.55            \\ \hline
3D DCT (4$\times$4$\times$4)     & $T=64$             &   0.61/1.11        &   31.56/33.21             \\
3D DCT (8$\times$8$\times$8)    & $T=512$             &  0.68/1.05         &  33.14/33.53            \\
3D DCT (16$\times$16$\times$16)* & $T=4096$             &  0.65/1.09         &  33.17/33.51             \\ \hline
\end{tabular}
}
\caption{\textbf{Ablation study: the impact of varying DCT block size} on the \textit{Ficus} scene in the Synthetic NeRF dataset (low/high bitrates), where * indicates the default configuration. These experimental configurations are only for matrices $\mathbf{M}$.}
\label{tab:tab3}
\end{table}

{\textbf{Effect of the Regularization Term.} This section analyzes how the $\mathcal{L}_{r}$ term affects compression performance. This term encourages the transformed coefficient values to become zero, which leads to a significant reduction in size when combined with entropy minimization. However, the bitrate is highly sensitive to the weighting parameters $\lambda_{r}$ and $\lambda_{e}$, making it difficult to analyze their effect. To avoid exhaustive search, we introduce a offset $\alpha$, which automatically determines $\lambda_{r}$ as $\lambda_{r}=\alpha\lambda_{e}$. As shown in Fig. \ref{fig:fig5}, it can be observed that adding the $\mathcal{L}_{r}$ term improves compression performance. When the offset $\alpha$ is close to 1 ($\lambda_{r}=\lambda_{e}$), it performs more effectively. However, when the value of $\alpha$ is too large or too small, it can negatively impact compression performance. In addition, in Fig. \ref{fig:fig5}, we illustrate the performance of entropy minimization in the spatial feature-grid domain instead of the frequency domain. It clearly demonstrates the advantages of entropy minimization in the frequency domain.

\begin{figure}[t]
\centering
          \begin{subfigure}{\linewidth}
         \centering
         \includegraphics[width=\linewidth, height=6.45cm]{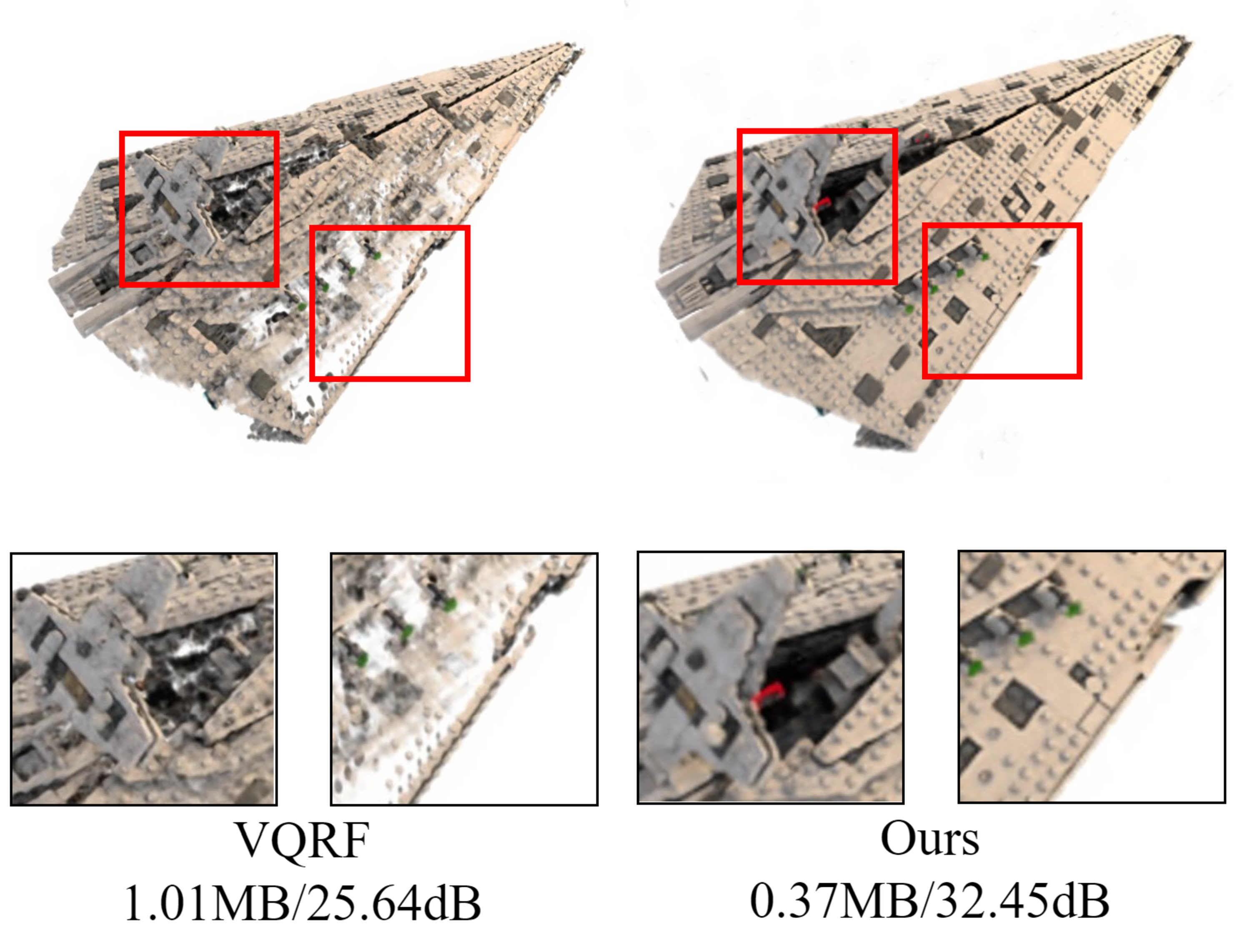}
     \end{subfigure}
     \caption{\textbf{Compression artifacts} of VQRF \cite{li2023compressing} and our proposed model ($\lambda_{e}=2e^{-10}$) for \textit{Destroyer} scene. VQRF is a pruning-based compression model.}
     \label{fig:fig6}
\end{figure}

\textbf{Impact of DCT Block Size.} The impact of DCT size is another important aspect of the proposed model. The entropy and regularization term involves applying the block-wise DCT to the feature-grid. Therefore, it is conceivable that the size of the DCT block will impact the performance. This ablation study focuses on the \textit{Ficus} scene, which shows clear trends for DCT block size. Tab. \ref{tab:tab3} shows the performance according to the DCT block size with matrices $\mathbf{M}$. It can be observed that a smaller total block size ($T<512$) has a significantly negative impact on performance. However, if the total block size $T$ is bigger than 512, there is no meaningful improvement regardless of the dimension. For vectors $\mathbf{v}$, the size of the DCT block does not have a significant impact. In addition, the size of the vectors is limited to applying a larger block, so we fixed 2D DCT 8$\times$8 for vectors $\mathbf{v}$. 


{\textbf{Compression Artifacts.} Fig. \ref{fig:fig6} illustrates the reconstruction errors that occur at extremely low bitrates. For the proposed model, the optimized tensor signal only contains low-frequency components of the DCT, which leads to block artifacts in the feature-grid. The reconstructed scene suffers from a loss of high-frequency information. However, using a pruning model like VQRF for compression may result in the loss of spatial information. Our proposed model has the advantage of reducing size while preserving visual quality even at extremely low bitrates. 

{\textbf{Compression Pipeline.} While the feature-grid is optimized after training, sparse representations themselves do not decrease the total size. Our design follows the compression pipeline of VQRF, but the compression target of our model is the transformed coefficients. The experimental results with the compression pipeline are listed in Tab. \ref{tab:tab4}. The VQRF model uses pruning and joint fine-tuning techniques to eliminate unnecessary weights without significant quality loss. Then, they apply 8-bit quantization to the feature-grid, resulting in a size of approximately 5.4MB. Finally, entropy coding is applied to reduce the size of the data and pack it into a binary bitstream for transmission. In contrast, our proposed model directly minimizes the entropy with our loss function. After performing DCT and 8-bit quantization, applying entropy coding to the transformed coefficients leads to a significant reduction in size. Compared to the uncompressed baseline, our proposed model achieves a compression ratio of 28× with a negligible drop in PSNR of only 0.1 dB. The size of our proposed model is 0.8MB for higher $\lambda_{e}$, with a loss of 0.95dB. 

\textbf{Complexity Overhead.} The training time for the proposed model (VM-192) is calculated using an NVIDIA A100 with 40 GB of memory. The DCT operation slightly increases the training time, while the entropy calculation significantly increases it. Since the entropy loss is applied after the feature-grid is fixed in coarse-to-fine training, our proposed model typically takes about 4 to 5 minutes longer than the baseline. The rendering time of the compressed model is comparable to the original one, as it only requires entropy decoding and inverse DCT.

\begin{table}
  \centering
  \resizebox{0.5\textwidth}{!}{
\begin{tabular}{c|ccc}
\hline
Method                                     & Size (MB)↓                 & PSNR (dB)↑ & SSIM↑ \\ \hline
TensoRF-VM baseline, 30k                   & 71.3$\mathrm{MB}$          & 33.15      & 0.962 \\ \hline
VQRF (fine-tuned) \cite{li2023compressing}                 & 21.9$\mathrm{MB}$         & 32.98      & 0.960 \\
$+$8bit quantization                       & 5.47$\mathrm{MB}$          & 32.88      & 0.959 \\
$+$Entropy coding                          & \textbf{3.52}$\mathrm{\textbf{MB}}$          & 32.88      & 0.959 \\ \hline
Ours, 30k ($\lambda_{\text {e}}=1e^{-11}$) & 71.6$\mathrm{MB}$          & 33.12      & 0.961 \\
$+$DCT, 8bit quantization                       & 18.1$\mathrm{MB}$          & 33.05      & 0.960 \\
$+$Entropy coding                          & \textbf{2.51}$\mathrm{\textbf{MB}}$ & 33.05      & 0.960 \\ \hline
Ours, 30k ($\lambda_{\text {e}}=2.5e^{-9}$)  & 71.6$\mathrm{MB}$          & 32.45      & 0.957 \\
$+$DCT, 8bit quantization                       & 18.1$\mathrm{MB}$          & 32.20      & 0.956 \\
$+$Entropy coding                          & \textbf{0.80}$\mathrm{\textbf{MB}}$ & 32.20      & 0.956 \\ \hline
\end{tabular}
}
\caption{\textbf{Comparison of VQRF \cite{li2023compressing} and our proposed model with the compression pipeline} discussed in \ref{sec:sec4.3}. The experimental results are averaged across all scenes from the Synthetic NeRF dataset.}
\label{tab:tab4}
\end{table}

\section{Conclusion}
\label{sec:conclusion}


We introduce ECRF, a novel compression framework specifically designed for tensorial radiance fields, but it can be applied to any grid-based neural fields. Unlike traditional compression methods, ECRF's primary goal is to directly minimize the entropy of the compression target. This optimization process takes place in the DCT coefficient domain, leading to more sparse representations for efficient compression. This is achieved through the utilization of a frequency-domain entropy parameterization, which is integrated into a compression pipeline that includes 8-bit quantization and entropy coding. Our proposed model effectively reduces the model size without compromising rendering quality, as demonstrated by experimental results showing superior performance, especially at low bitrates.

\clearpage

{
    \small
    \bibliographystyle{ieeenat_fullname}
    \bibliography{main}
}

\end{document}